\documentclass{article}
\usepackage{spconf,amsmath,graphicx}
\usepackage[pagebackref,breaklinks,colorlinks]{hyperref}
\usepackage{algorithm}
\usepackage{algorithmic}
\usepackage{diagbox}
\usepackage{graphicx}
\usepackage{booktabs}
\usepackage{comment}
\newcommand{\tenpt}[1]{{\fontsize{10pt}{12pt}\selectfont #1}}
\usepackage{multirow}
\usepackage{subfig}
\usepackage{mathtools}
\usepackage{amsmath}
\usepackage{amsfonts}
\usepackage{amssymb}
\usepackage{afterpage}
\usepackage{pifont}

\title{Killing it with Zero-Shot: \\Adversarially Robust Novelty Detection}
\name{
    \begin{tabular}{c}
        Hossein Mirzaei\textsuperscript{1}, 
        Mohammad Jafari\textsuperscript{1}, 
        Hamid Reza Dehbashi\textsuperscript{1}, 
        Zeinab Sadat Taghavi\textsuperscript{1}, \\
        \textit{Mohammad Sabokrou}\textsuperscript{2}, 
        \textit{Mohammad Hossein Rohban}\textsuperscript{1}
    \end{tabular}
}

\address{
    \textsuperscript{1} Sharif University of Technology, Tehran, Iran \\
    \textsuperscript{2} Okinawa Institute of Science and Technology, Okinawa, Japan
}

\begin{document}

\maketitle
\begin{abstract}
 
Novelty Detection (ND) plays a crucial role in machine learning by identifying new or unseen data during model inference. This capability is especially important for the safe and reliable operation of automated systems. Despite advances in this field, existing techniques often fail to maintain their performance when subject to adversarial attacks. Our research addresses this gap by marrying the merits of nearest-neighbor algorithms with robust features obtained from models pretrained on ImageNet. We focus on enhancing the robustness and performance of ND algorithms. Experimental results demonstrate that our approach significantly outperforms current state-of-the-art methods across various benchmarks, particularly under adversarial conditions. By incorporating robust pretrained features into the k-NN algorithm, we establish a new standard for performance and robustness in the field of robust ND. This work opens up new avenues for research aimed at fortifying machine learning systems against adversarial vulnerabilities. Our implementation is publicly available at \url{https://github.com/rohban-lab/ZARND}.
\end{abstract} 
\section{Introduction}
\label{sec:intro}

\begin{figure}[t]
  \begin{center}
    \hspace{-0.5cm} \includegraphics[width=0.9\linewidth]{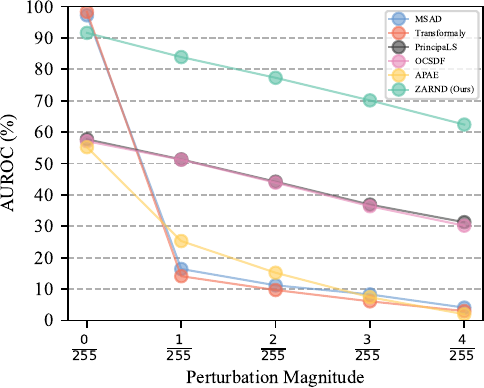}
    \caption{This plot evaluates ND methods—MSAD\cite{reiss2021mean}, Transformaly\cite{cohen2021transformaly}, PrincipaLS\cite{lo2022adversarially}, OCSDF\cite{bethune2023robust}, APAE\cite{goodge2021robustness}, and ZARND (Ours)—for their robustness, measured in AUROC (\%) on the CIFAR-10 dataset. It highlights ZARND's superior performance and emphasizes the need for more robust ND algorithms.}
    \label{fig:Motivation}
  \end{center}
\end{figure}

Anomaly detection plays an important role in computer vision, for example, in healthcare, industry, and autonomous driving \cite{perera2021one}. Its ability to identify new or unseen visual patterns during inference is essential for the safety and reliability of ML-based systems \cite{nguyen2015deep}. 
Recent ND methods have shown significant results in clean settings. However, their performance significantly deteriorates under adversarial attacks to the extent that it is worse than random guessing. For instance, the SOTA ND results on the MVTecAD dataset \cite{bergmann2019mvtec} achieve near-perfect performance, with 100\% AUROC (Area Under the Receiver Operating Characteristics) \cite{batzner2023efficientad}. However, when tiny adversarial perturbations are added to the data, the performance of all previous works drops to near 0\% AUROC. This emphasizes the necessity for further research to bridge the performance gap of novelty detection between clean and adversarial settings \cite{salehi2021unified}. Our main goal is to propose a method that not only performs well in benign settings but is also resilient to adversarial attacks that attempt to convert normal samples to anomalies and vice versa. 

Motivated by this, we introduce a framework named Zero-shot Adversarially Robust Novelty Detection (ZARND), which leverages an adversarially robust pre-trained model as a backbone and extracts features from the training data using the last layer of the backbone. During testing, input data is passed through the backbone, and anomaly scores are calculated by considering the features of the training data, utilizing well-established algorithms such as k-Nearest Neighbors (k-NN) \cite{Mucherino2009}.

While many previous methods utilize pre-trained models for ND as a downstream task, we are the first to employ an adversarially robust pre-trained model for ND task, as illustrated in Figure \ref{fig:Motivation}. The features extracted by our backbone are both descriptive and resilient against adversarial manipulations, enabling us to achieve competitive results in clean settings and establish a SOTA performance in adversarial settings. We summarize the main contributions of this paper as follows:

\begin{itemize}
    \item Our proposed zero-shot ND method, ZARND, achieves significant results on various datasets, including medical and tiny datasets, highlighting the applicability of our work to real-world applications.
    

    \item We extensively compared our method in unsupervised ND settings, including one-class and unlabeled multi-class scenarios, with both clean and adversarially trained proposed methods, surpassing their performance by a margin of up to 40\% AUROC.
\end{itemize}

\section{Background \& Related Work}
\label{sec:background}

\textbf{ND.}
Various self-supervised approaches have been introduced in ND to learn the normal sample distribution.
DeepSVDDs utilizes an autoencoder to learn distribution features, while CSI\cite{tack2020csi} employs a contrastive loss for distribution learning. 
Moreover FITYMI \cite{mirzaei2022fake}, DN2\cite{bergman2020deep}, PANDA\cite{reiss2021panda}, Transformaly \cite{cohen2021transformaly} and MSAD\cite{reiss2021mean} leverage pre-trained models for gaining insights. Additionally, some methods have ventured into ND in an adversarial context; APAE\cite{goodge2021robustness} proposes using approximate projection and feature weighting for enhancing adversarial robustness, and PrincipaLS\cite{lo2022adversarially} suggests employing Principal Latent Space as a defense strategy for adversarially robust ND. 
OCSDF\cite{bethune2023robust} focuses on robustness by learning a signed distance function, interpreted as a normality score. The OSAD\cite{shao2022open} method integrates adversarial training loss with an auto-encoder loss aimed at reconstructing the original image from its adversarial counterpart. RODEO \cite{mirzaeirodeo} and AROS \cite{mirzaei2024adversarially} are also recent adversarially robust detection methods \cite{mirzaei2024universal,mirzaeiscanning,moakhar2023seeking,taghavi2023change,rahimi-etal-2024-hallusafe,taghavi2023imaginations,taghavi-etal-2023-ebhaam,taghavi2024backdooring,rahimi-etal-2024-nimz,ebrahimi2024sharif,ebrahimi2024sharifa}.\\
\textbf{Adversarial Attacks.}
An adversarial attack involves maliciously perturbing a data point \( \mathbf{x} \) with an associated label \( y \) to generate a new point \( \mathbf{x^*} \) that maximizes the loss function \( \ell(\mathbf{x^*}; y) \). 
In this setting, \( \mathbf{x^*} \) is an adversarial example of \( \mathbf{x} \), and the difference \( (\mathbf{x^*} - \mathbf{x}) \) is termed as adversarial noise. An upper bound \( \epsilon \) limits the \( l_p \) norm of the adversarial noise to avoid altering the semantics of the original data. 
Specifically, an adversarial example \( \mathbf{x^*} \) must satisfy: 
\[
\quad\quad \mathbf{x^*} = \arg \max_{\mathbf{x^{\prime}}} \ell(\mathbf{x^{\prime}}; y), \quad \quad \|\mathbf{x}-\mathbf{x^*}\|_p \leq \epsilon
\]
The well-known projected Gradient Descent (PGD) technique \cite{madry2017towards} iteratively maximizes the loss function by updating \( \mathbf{x^*} \) in the direction of the gradient \( \nabla_{\mathbf{x}} \ell(\mathbf{x^*}; y) \). 
During each iteration, the adversarial noise is constrained within an \( \ell_{\infty} \)-ball of radius \( \epsilon \):
\[\quad   \mathbf{x_0^*}=\mathbf{x}, \quad \mathbf{x_{t+1}^*}=\mathbf{x_t^*}+\alpha. \operatorname{sign}\left(\nabla_{\mathbf{x}} \ell\left(\mathbf{x_t^*}, y\right)\right)\]
\[
\ \ \ \mathbf{x^*} = \mathbf{x^{*}_T}
\]

\section{Problem Statement}
\label{sec:problem-statement}


\textbf{Setup.}
In ND, our focus is on a specialized training set that comprises solely normal, unlabeled samples. This training set is formally denoted as:

\[
\mathcal{X}_{\text{train}} = \{\mathbf{x}_1, \mathbf{x}_2, \ldots, \mathbf{x}_n\}
\]
Here, \( \mathcal{X}_{\text{train}} \) represents the collection of all 'normal' samples, and each \( \mathbf{x}_i \) is an individual sample. These samples are elements of a broader feature space, denoted as \( \mathcal{X} \). Our main goal is to develop a function \( f \) that maps from this feature space to a binary output, which can be formally defined as:

\[
f: \mathcal{X} \rightarrow \{0, 1\}
\]
In the context of this function:
\begin{itemize}
    \item \( f(\mathbf{x}) = 0 \) implies that the sample \( \mathbf{x} \) is 'normal'
    \item \( f(\mathbf{x}) = 1 \) implies that the sample \( \mathbf{x} \) is an 'outlier'
\end{itemize}

\noindent\textbf{Challenges.} One of the main difficulties in ND comes from the unsupervised nature of the training set \( \mathcal{X}_{\text{train}} \), which lacks labels. Algorithms have the task of autonomously defining what is 'normal,' based solely on \( \mathcal{X}_{\text{train}} \)'s data distribution. Another challenge arises from the common closed-set assumption in machine learning, which stipulates that test data should come from the same distribution as training data. In ND, this is not often the case, complicating the identification of 'novel' samples.

ND's challenges extend to the quality of training data and the absence of labels. The training set's unlabeled nature necessitates that ND algorithms identify what is genuinely 'normal' autonomously. Noise in the dataset places importance on choosing an effective distance metric. Simple metrics, such as pixel-level spaces, are inadequate as they fail to capture the underlying semantic differences that separate normal from novel instances. This limitation is accentuated when deep neural networks are used to transform the data. In this scenario, selecting a suitable threshold for differentiation becomes even more challenging, justifying the common use of AUROC as an evaluation metric.

\section{Method}
\label{sec:method}

\begin{figure}[t]
  \begin{center}
    \includegraphics[width=\linewidth]{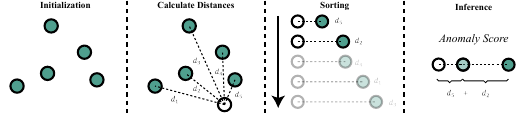}
    \caption{The four key steps of our ND framework are illustrated. (a) \textit{Initialization}: Extracting features from the normal set. (b) \textit{Calculate Distance}: Computing the Euclidean difference between the test sample and features from the normal set. (c) \textit{Sorting}: Ordering the distances from smallest to largest. (d) \textit{Anomaly Score}: Summing up the first \( K \) smallest distances to generate the anomaly score, where \( K = 2 \).}

    \label{fig:Method}
  \end{center}
\end{figure}

\subsection{Initialization: Feature Extraction}
\label{subsec:initialization_feature_extraction}

Our methodology begins with the critical step of feature extraction, setting the stage for all downstream operations. The primary objective at this stage is to generate a comprehensive and highly expressive feature space that is not only adaptable but also generalizable across various types of data sets. To this end, we employ pre-trained ResNet architectures as the backbone of our feature extraction mechanism, denoted by the function \( h \).

Given an input sample \( \mathbf{x}_i \), the feature extraction function \( h \) performs a mapping operation that transforms \( \mathbf{x}_i \) into a high-dimensional feature vector \( \mathbf{z}_i = h(\mathbf{x}_i) \). By leveraging the extensive training and architectural benefits of ResNet, our feature extraction phase is capable of capturing complex, high-level semantics from the data. 

Notably, our methodology deviates from conventional approaches by discarding the terminal classification layer commonly found in neural network architectures. Instead, we focus on an intermediary feature map \( \mathbf{F} \) as the end product of this phase. This decision is grounded in our goal to construct a feature space that is domain-agnostic, optimized specifically for the task of anomaly detection. By doing so, we ensure that the resulting feature space can capture the intricate data structures and patterns needed for high-quality novelty detection across diverse application domains.

\subsection{Calculate Distance: Computing Euclidean Distances}
\label{subsec:calculate_distance}

After initializing our feature space through the extraction process, the next crucial step in our methodology is to compute the Euclidean distances between a given test sample and the features stored in our feature bank. This is a pivotal operation as it lays the foundation for the anomaly score, which ultimately determines whether a data point is an outlier or not.

To carry out this operation, we introduce a test sample \( \mathbf{z}_{\text{test}} \) and compare it with our feature bank \( \mathbf{z} \), which is a collection of high-dimensional vectors extracted from in-distribution or 'normal' samples. The objective is to quantify the distance between \( \mathbf{z}_{\text{test}} \) and each feature vector \( \mathbf{z}_i \) in the feature bank.

Formally, the Euclidean distance between \( \mathbf{z}_{\text{test}} \) and \( \mathbf{z}_i \) is calculated as follows:
\[
d(\mathbf{z}_{\text{test}}, \mathbf{z}_i) = || \mathbf{z}_{\text{test}} - \mathbf{z}_i ||_2
\]

The \( L_2 \) norm is employed here as our distance metric, in accordance with the widely-accepted practice in anomaly detection tasks. By computing these distances, we set the stage for the subsequent sorting operation, which aids in the formulation of a precise anomaly score.

\subsection{Sorting: Ordering Euclidean Distances}
\label{subsec:sorting}

Once the Euclidean distances have been computed between the test sample \( \mathbf{z}_{\text{test}} \) and all the feature vectors \( \mathbf{z}_i \) in our feature bank, the next phase is to sort these distances. Sorting is crucial as it enables us to identify which vectors in the feature bank are closest to \( \mathbf{z}_{\text{test}} \), and therefore most relevant for the anomaly scoring process.

Let \( D = \{ d_1, d_2, \ldots, d_N \} \) be the set of Euclidean distances, where \( N \) is the total number of feature vectors in our feature bank. The sorting operation can be formally described as: \text{Sorted Distances} = \text{sort}(D). This orders the distances in ascending order, preparing us for the final step: the calculation of the anomaly score using the k-nearest neighbors from this sorted list.

\subsection{Anomaly Score: Calculating k-NN-Based Score}
\label{subsec:anomaly_score}

The culmination of our methodology is the computation of the anomaly score, which is a decisive metric for categorizing a data point as normal or an outlier. We employ the k-NN algorithm to accomplish this, taking advantage of its efficacy in capturing local data characteristics without making assumptions about the global data distribution. To clarify the formula, let \( D_{\text{sorted}} = \{ d_1, d_2, \ldots, d_k \} \) be the set of k-smallest distances obtained from the sorted list in the previous step. The anomaly score \( S \) for a test sample \( \mathbf{z}_{\text{test}} \) can be formally defined as: $\quad S = \sum_{d \in D_{\text{sorted}}} d$. Here, \( S \) is the sum of the k-smallest Euclidean distances between the test feature vector \( \mathbf{z}_{\text{test}} \) and the k-nearest feature vectors from our feature bank. A higher value of \( S \) indicates that the test sample is more likely to be an outlier, making \( S \) a critical metric for our Novelty Detection framework. Our proposed pipeline is illustrated in Figure \ref{fig:Method}.

\section{Experimental Results}

\begin{table*}[thb]
    \centering
    \resizebox{0.95\linewidth}{!}{
    \ninept
    \setlength{\tabcolsep}{4pt} 
    \begin{tabular}{lll*{14}{c}} 
    \toprule
     \noalign{\smallskip}
     \multicolumn{2}{l}{\multirow{5}{*}{\textbf{\tenpt{Dataset}}}}& \multirow{5}{*}{\rotatebox[origin=c]{90}{\textbf{\tenpt{Attack}}}} & \multicolumn{12}{c}{\textbf{\tenpt{Method}}} \\
     \noalign{\smallskip}
    \cmidrule(lr){4-15}
    \noalign{\smallskip}
    & &  & \multirow{4}{*}{DeepSVDD} &  \multirow{4}{*}{MSAD} &  \multirow{4}{*}{Transformaly} &  \multirow{4}{*}{PatchCore} &  \multirow{4}{*}{PrincipaLS} &  \multirow{4}{*}{OCSDF} &  \multirow{4}{*}{APAE} &   \multicolumn{5}{c}{\tenpt{Zero-Shot}} \\
    \noalign{\smallskip}
    \cmidrule(lr){11-15}
    & & & & & & & & & &  \multicolumn{2}{c}{Clean Pre-Trained} &  &  \multicolumn{2}{c}{Robust Pre-Trained}  \\
    \cmidrule(lr){11-12} \cmidrule(lr){13-15}
     & & & & & & & & & & GMM & k-NN & & GMM & k-NN (\textbf{ZARND}) \\
    \noalign{\smallskip}
    \specialrule{0.8pt}{\aboverulesep}{\belowrulesep}

    \multirow{8}{*}[-0.3cm]{\rotatebox[origin=c]{90}{\textbf{\tenpt{Low-Res}}}}
    &  \multirow{2}{*}{CIFAR-10} & 
    
    Clean & 64.8 & \underline{97.2} & \textbf{98.3} & 68.3 & 57.7 & 57.1 & 55.2 & 92.24 & 93.9 & & 87.3 & 89.7\\
    
    & & PGD/AA & 31.7 / 16.6 & 4.8 / 0.7 & 3.7 / 2.4 & 3.9 / 2.5 & 33.2 /
    29.6 & 31.3 / 26.9 & 2.2 / 2.0 & 0.0 / 0.0 & 0.0 / 0.0 & & \textbf{62.7} / \underline{60.3} & \underline{62.4} / \textbf{60.6}\\
    
    \cmidrule(lr){2-15}  
    
    &  \multirow{2}{*}{CIFAR-100} &
    
    Clean & 67.0 & \underline{96.4} & \textbf{97.3} & 66.8 & 52.0 & 48.2 & 51.8 & 89.7 & 93.6 &  & 83.0 & 88.4 \\
    
    & &PGD/AA & 23.7 / 14.3 & 8.4 / 10.7 & 9.4 / 7.3 & 4.3 / 3.5 & 26.2 / 24.6 & 23.5 / 19.8 & 4.1 / 0.9 & 0.0 / 0.0 & 0.0 / 0.0 &  & \underline{53.2} / \underline{50.7} &  \textbf{62.9} / \textbf{60.1} \\
    
    \cmidrule(lr){2-15}  
    
    &  \multirow{2}{*}{MNIST} &
    
    Clean & 94.8 & 96.0 & 94.8 & 83.2 & \underline{97.3} & 95.5 & 92.5 & 91.3 & 96.0 & & 95.0 & \textbf{99.0}\\
    
    & &PGD/AA & 18.8 / 15.4 & 3.2 / 14.1 & 18.9 / 11.6 & 2.6 / 2.4 & 83.1 / 77.3 & 68.9 / 66.2 & 34.7 / 28.6 & 0.0 / 0.0 & 0.0 / 0.0 & & \underline{90.5} / \underline{84.1}& \textbf{97.1} / \textbf{95.2}\\
    
   \cmidrule(lr){2-15}   
    
    &  \multirow{2}{*}{FMNIST} &
    
    Clean & {94.5} & 94.2 & {94.4} & 77.4 & 91.0 & 90.6 & 86.1 & 88.4 & \underline{94.7} & & 92.9 & \textbf{95.0} \\
    & &PGD/AA & 57.9 / 48.1 & 6.6 / 3.7 & 7.4 / 2.8 & 5.5 / 2.8 & 69.2 / 67.5 & 64.9 / 59.6 & 19.5 / 13.3 & 0.0 / 0.0 & 0.0 / 0.0 & & \underline{87.4} / \underline{81.5}& \textbf{89.5} / \textbf{85.9}\\
    
    \specialrule{0.8pt}{\aboverulesep}{\belowrulesep}
    
    \multirow{10}{*}[-0.3cm]{\rotatebox[origin=c]{90}{\textbf{\tenpt{High-Res}}}}
    
    &  \multirow{2}{*}{MVTecAD} &
    
    Clean &  67.0 & {87.2} & \underline{87.9} &\textbf{ 99.6} & 63.8 & 58.7 & 62.1 & 78.7 & 74.8 & & 74.2 & 71.6 \\
    
    & &PGD/AA & 2.6 / 0.0 & 0.9 / 0.0 & 0.0 / 0.0 & 7.2 / 4.8 & {24.3} / {12.6} & 5.2 / 0.3 & 4.7 / 1.8 & 0.0 / 0.0 & 0.0 / 0.0 & & \textbf{38.5} / \textbf{32.0} & \underline{30.1} / \underline{28.8}\\
    
    \cmidrule(lr){2-15}  
    
    &  \multirow{2}{*}{Head-CT} &
    
    Clean & 62.5 & 59.4 & 78.1 & \textbf{98.5} & 68.9 & 62.4 & 68.1 & 80.6 & \underline{90.9} & & 75.6 & 89.0\\
    
    & &PGD/AA & 0.0 / 0.0 & 0.0 / 0.0 & 6.4 / 3.2 & 1.5 / 0.0 & 27.8 / 16.2 & 13.1 / 8.5 & 6.6 / 3.8 & 0.0 / 0.0 & 0.0/ 0.0 & & \underline{66.7} / \underline{62.8} & \textbf{80.1} / \textbf{75.6}\\

    \cmidrule(lr){2-15}  
    
    &  \multirow{2}{*}{BrainMRI} &
    
    Clean & 74.5 & \textbf{99.9} & \underline{98.3} & 91.4 & 70.2 & 63.2 & 55.4 & 69.0 & 82.3 & & 62.3 & 76.8\\
    
    & &PGD/AA & 4.3 / 2.1 & 1.7 / 0.0 & 5.2 / 1.6 & 0.0 / 0.4 & 33.5 / 17.8 & 20.4 / 12.5 & 9.7 / 8.3 & 0.0 / 0.0 & 0.0 / 0.0 & & \underline{47.6} / \underline{44.9} & \textbf{69.6} / \textbf{65.7}\\
    
    \cmidrule(lr){2-15}  
    
    &  \multirow{2}{*}{Tumor} &
    
    Clean & 70.8 & {95.1} & {97.4} & 92.8 & 73.5 & 65.2 & 64.6 & 88.6 & \textbf{99.9} & & 80.8 & \underline{98.9}\\
    
    & &PGD/AA & 1.7 / 0.0 & 0.1 / 0.0 & 7.4 / 5.1 & 9.3 / 6.1 & 25.2 / 14.7 & 17.9 / 10.1 & 15.8 / 8.3 & 0.0 / 0.0 & 16.4 / 14.8 & & \underline{71.8} / \underline{65.0} & \textbf{82.7} / \textbf{78.6} \\
    
    \cmidrule(lr){2-15}  
    
    &  \multirow{2}{*}{Covid-19} &
    
    Clean & 61.9 & {89.2} & {91.0} & 77.7 & 54.2 & 46.1 & 50.7 & \underline{91.6} & \textbf{92.2} & & 74.6 & 72.1\\
    
    & &PGD/AA & 0.0 / 0.0& 4.7 / 1.9 & 10.6 / 4.4 & 4.2 / 0.5 & 15.3 / 9.1 & 9.0 / 6.5 & 11.2 / 8.7 & 0.0 / 0.0 & 19.8/ 14.1 & & \underline{49.3} / \underline{40.4} & \textbf{56.7} / \textbf{54.8} \\
    
    \specialrule{0.8pt}{\aboverulesep}{\belowrulesep}
    
    \end{tabular}
    }
    \caption{In this table, we assess ND methods on various datasets against PGD-100, AutoAttack (AA), and a clean setting, using the AUROC(\%) metric. We use Gaussian Mixture Model (GMM) and k-NN as the anomaly scores for the Zero-Shot setting. ZARND, our introduced method, outperforms the other methods in the adversarial setting according to the table. Perturbations: $ \epsilon = \frac{4}{255} $ for low-res and $ \epsilon = \frac{2}{255} $ for high-res datasets. Best scores are \textbf{bolded}; second-bests are \underline{underlined}.}

    \label{Table1:ND}
\end{table*}

\begin{table}[thb]
    \centering
    \resizebox{\linewidth}{!}{
    \ninept
    \setlength{\tabcolsep}{4pt} 
    \begin{tabular}{lll*{8}{c}} 
    \toprule
     \noalign{\smallskip}
     \multicolumn{1}{c}{\multirow{3}{*}{\textbf{\tenpt{In}}}}& \multirow{3}{*}{\textbf{\tenpt{Out}}} & \multicolumn{8}{c}{\textbf{\tenpt{Method}}} \\
     \noalign{\smallskip}
    \cmidrule(lr){3-10}
    & & \multicolumn{2}{c}{\tenpt{PrincipaLS}} &  \multicolumn{2}{c}{\tenpt{OCSDF}} &  \multicolumn{2}{c}{\tenpt{APAE}} &   \multicolumn{2}{c}{\tenpt{\textbf{ZARND (Ours)}}} \\
    \cmidrule(lr){3-4} \cmidrule(lr){5-6} \cmidrule(lr){7-8} \cmidrule(lr){9-10}
     & & Clean & PGD-100 & Clean & PGD-100 & Clean & PGD-100 & Clean & PGD-100  \\
    \specialrule{0.8pt}{\aboverulesep}{\belowrulesep}

    \multirow{4}{*}{\rotatebox[origin=c]{90}{\textbf{CIFAR-10}}}
    &  CIFAR-100 & {54.8} & {14.6} & 51.0 & 12.8 & 53.6 & 1.2 & \underline{76.6} & \textbf{40.2} \\
    &  SVHN & {72.1} & {23.6} & 67.7 & 18.8 & 60.8 & 2.1 & \underline{84.3} & \textbf{48.8} \\
    &  MNIST & {82.5} & {42.7} & 74.2 & 37.4 &‌ 71.3 & 15.3 & \underline{99.4} & \textbf{95.2}\\
    &  FMNIST & {78.3} & {38.5} & 64.5 & 33.7 & 59.4 & 9.4 & \underline{98.2} & \textbf{89.4}\\
    
    \midrule
    
    
    \multirow{4}{*}{\rotatebox[origin=c]{90}{\textbf{CIFAR-100}}}
    &  CIFAR-10 & 47.6 & {8.1} & {51.1} & 6.3 & 50.5 & 0.7 & \underline{64.6} &‌ \textbf{26.8} \\
    &  SVHN & {66.3} & {13.2} & 58.7 & 9.2 & 58.1 & 1.1 & \underline{70.0} &‌ \textbf{32.7} \\
    &  MNIST & {80.4} & {30.4} & 76.4 & 28.9 &‌ 74.7 & 11.8 & \underline{87.0} &‌ \textbf{61.7} \\
    &  FMNIST & {72.7} & {18.7} & 62.8 & 14.3 & 60.9 & 9.7 & \underline{97.3‌} &‌ \textbf{85.8} \\
    \noalign{\smallskip}

    \bottomrule
    
    \end{tabular}
    }
    \caption{AUROC (\%) of various ND methods trained on unlabeled CIFAR-10 and CIFAR-100. According to the table, ZARND performs much better than the other methods. Best adversarial scores are \textbf{bolded}; best clean scores are \underline{underlined}.}
    \label{Table2:UOOD Detection}
\end{table}
In this section, we conduct comprehensive experiments to assess various outlier detection methods, encompassing both the clean and adversarially trained approaches and our method in the context of adversarial attacks.\\
\textbf{Setup.} We utilize the ResNet-18 architecture as the backbone of our neural network and incorporate pre-trained weights from \cite{DBLP:journals/corr/abs-2007-08489}. \\
\textbf{Evaluation Attack.} We set the value of $\epsilon$ to $\frac{4}{255}$ for low-resolution datasets and $\frac{2}{255}$ for high-resolution ones. For the models' evaluation, we use a single random restart for the attack, with random initialization within the range of $(-\epsilon, \epsilon)$ , and perform 100 steps. Furthermore, we select the attack step size as $\alpha=2.5 \times \frac{\epsilon}{N}$.
In addition to the PGD attack, we have evaluated models using AutoAttack \cite{croce2020reliable}, a recently introduced method designed to enhance robustness evaluation by implementing more potent attacks.\\
We introduce a label-dependent function \( \beta(y) \) to design effective adversarial perturbations:

\[
\beta(y) = 
\begin{cases} 
1, & \text{if } y = 0 \\
-1, & \text{if } y = 1 
\end{cases}
\]
Our methodology incorporates adversarial perturbations targeting the raw input images \( \mathbf{x} \) instead of their corresponding feature vectors \( \mathbf{z} \). We use the PGD to iteratively perturb each image. The perturbation is designed to affect the k-NN-based anomaly score as follows directly:
\[\mathbf{x}^*_{t+1} = \mathbf{x}^*_{t} + \alpha \beta(y) \text{ sign}(\nabla_{\mathbf{x}^*_t} S(h(\mathbf{x}^*_t)))\]
Here, \( \alpha \) specifies the step size, \( t \) the iteration number, and \( h(\mathbf{x}^*_t) \) the feature vector derived from the perturbed image.

\noindent\textbf{Datasets and metrics.} We included CIFAR-10, CIFAR-100, MNIST and FashionMNIST for the low-resolution datasets. Furthermore, we performed experiments on medical and industrial high-resolution datasets, namely Head-CT\cite{felipe-campos-kitamura_2018}, MVTec-ad, Brain-MRI\cite{brainmri}, Covid-19\cite{cohen2020covid}, and Tumor Detection\cite{msoud-nickparvar_2021}. We reported the results in Table \ref{Table1:ND}. We use AUROC as a well-known classification criterion. The AUROC value is in the range [0, 1], and the closer it is to 1, the better the classifier performance.

\noindent\textbf{ND.}
We run experiments across \( N \) classes, treating each class as normal in turn while all other classes act as outliers. Averaged results across all \( N \) setups are reported in Table \ref{Table1:ND}.

\noindent \textbf{Unlabeled Out-Of-Distribution (OOD) Detection.}
We introduce a label-free approach to OOD detection, using two distinct datasets for in-distribution and OOD samples. This binary problem could significantly benefit ND. Relevant findings are highlighted in Table \ref{Table2:UOOD Detection}.

\section{Conclusion}
\label{sec:conclusion}
Our work introduces a robust and practical approach to image-based ND, leveraging k-NN and feature extraction. Our method performs well in AUROC metrics and strongly resists adversarial attacks. We aim to further enhance the robustness and adaptability of our system to diverse applications.



\bibliographystyle{IEEEbib}
\bibliography{killing_with_zero_shot}

\begin{thebibliography}{10}

\bibitem{reiss2021mean}
Tal Reiss and Yedid Hoshen,
\newblock ``Mean-shifted contrastive loss for anomaly detection,''
\newblock {\em arXiv preprint arXiv:2106.03844}, 2021.

\bibitem{cohen2021transformaly}
Matan~Jacob Cohen and Shai Avidan,
\newblock ``Transformaly--two (feature spaces) are better than one,''
\newblock {\em arXiv preprint arXiv:2112.04185}, 2021.

\bibitem{lo2022adversarially}
Shao-Yuan Lo, Poojan Oza, and Vishal~M Patel,
\newblock ``Adversarially robust one-class novelty detection,''
\newblock {\em IEEE Transactions on Pattern Analysis and Machine Intelligence}, 2022.

\bibitem{bethune2023robust}
Louis B{\'e}thune, Paul Novello, Thibaut Boissin, Guillaume Coiffier, Mathieu Serrurier, Quentin Vincenot, and Andres Troya-Galvis,
\newblock ``Robust one-class classification with signed distance function using 1-lipschitz neural networks,''
\newblock {\em arXiv preprint arXiv:2303.01978}, 2023.

\bibitem{goodge2021robustness}
Adam Goodge, Bryan Hooi, See~Kiong Ng, and Wee~Siong Ng,
\newblock ``Robustness of autoencoders for anomaly detection under adversarial impact,''
\newblock in {\em Proceedings of the Twenty-Ninth International Conference on International Joint Conferences on Artificial Intelligence}, 2021, pp. 1244--1250.

\bibitem{perera2021one}
Pramuditha Perera, Poojan Oza, and Vishal~M Patel,
\newblock ``One-class classification: A survey,''
\newblock {\em arXiv preprint arXiv:2101.03064}, 2021.

\bibitem{nguyen2015deep}
Anh Nguyen, Jason Yosinski, and Jeff Clune,
\newblock ``Deep neural networks are easily fooled: High confidence predictions for unrecognizable images,''
\newblock in {\em Proceedings of the IEEE conference on computer vision and pattern recognition}, 2015, pp. 427--436.

\bibitem{bergmann2019mvtec}
Paul Bergmann, Michael Fauser, David Sattlegger, and Carsten Steger,
\newblock ``Mvtec ad--a comprehensive real-world dataset for unsupervised anomaly detection,''
\newblock in {\em Proceedings of the IEEE/CVF conference on computer vision and pattern recognition}, 2019, pp. 9592--9600.

\bibitem{batzner2023efficientad}
Kilian Batzner, Lars Heckler, and Rebecca K{\"o}nig,
\newblock ``Efficientad: Accurate visual anomaly detection at millisecond-level latencies,''
\newblock {\em arXiv preprint arXiv:2303.14535}, 2023.

\bibitem{salehi2021unified}
Mohammadreza Salehi, Hossein Mirzaei, Dan Hendrycks, Yixuan Li, Mohammad~Hossein Rohban, and Mohammad Sabokrou,
\newblock ``A unified survey on anomaly, novelty, open-set, and out-of-distribution detection: Solutions and future challenges,''
\newblock {\em arXiv preprint arXiv:2110.14051}, 2021.

\bibitem{Mucherino2009}
Antonio Mucherino, Petraq~J. Papajorgji, and Panos~M. Pardalos,
\newblock {\em k-Nearest Neighbor Classification}, pp. 83--106,
\newblock Springer New York, New York, NY, 2009.

\bibitem{tack2020csi}
Jihoon Tack, Sangwoo Mo, Jongheon Jeong, and Jinwoo Shin,
\newblock ``Csi: Novelty detection via contrastive learning on distributionally shifted instances,''
\newblock {\em Advances in neural information processing systems}, vol. 33, pp. 11839--11852, 2020.

\bibitem{mirzaei2022fake}
Hossein Mirzaei, Mohammadreza Salehi, Sajjad Shahabi, Efstratios Gavves, Cees~GM Snoek, Mohammad Sabokrou, and Mohammad~Hossein Rohban,
\newblock ``Fake it till you make it: Near-distribution novelty detection by score-based generative models,''
\newblock {\em arXiv preprint arXiv:2205.14297}, 2022.

\bibitem{bergman2020deep}
Liron Bergman, Niv Cohen, and Yedid Hoshen,
\newblock ``Deep nearest neighbor anomaly detection,''
\newblock {\em arXiv preprint arXiv:2002.10445}, 2020.

\bibitem{reiss2021panda}
Tal Reiss, Niv Cohen, Liron Bergman, and Yedid Hoshen,
\newblock ``Panda: Adapting pretrained features for anomaly detection and segmentation,''
\newblock in {\em Proceedings of the IEEE/CVF Conference on Computer Vision and Pattern Recognition}, 2021, pp. 2806--2814.

\bibitem{shao2022open}
Rui Shao, Pramuditha Perera, Pong~C Yuen, and Vishal~M Patel,
\newblock ``Open-set adversarial defense with clean-adversarial mutual learning,''
\newblock {\em International Journal of Computer Vision}, vol. 130, no. 4, pp. 1070--1087, 2022.

\bibitem{mirzaeirodeo}
Hossein Mirzaei, Mohammad Jafari, Hamid~Reza Dehbashi, Ali Ansari, Sepehr Ghobadi, Masoud Hadi, Arshia~Soltani Moakhar, Mohammad Azizmalayeri, Mahdieh~Soleymani Baghshah, and Mohammad~Hossein Rohban,
\newblock ``Rodeo: Robust outlier detection via exposing adaptive out-of-distribution samples,''
\newblock in {\em Forty-first International Conference on Machine Learning}, 2024.

\bibitem{mirzaei2024adversarially}
Hossein Mirzaei and Mackenzie~W Mathis,
\newblock ``Adversarially robust out-of-distribution detection using lyapunov-stabilized embeddings,''
\newblock {\em arXiv preprint arXiv:2410.10744}, 2024.

\bibitem{mirzaei2024universal}
Hossein Mirzaei, Mojtaba Nafez, Mohammad Jafari, Mohammad~Bagher Soltani, Mohammad Azizmalayeri, Jafar Habibi, Mohammad Sabokrou, and Mohammad~Hossein Rohban,
\newblock ``Universal novelty detection through adaptive contrastive learning,''
\newblock in {\em Proceedings of the IEEE/CVF Conference on Computer Vision and Pattern Recognition}, 2024, pp. 22914--22923.

\bibitem{mirzaeiscanning}
Hossein Mirzaei, Ali Ansari, Bahar~Dibaei Nia, Mojtaba Nafez, Moein Madadi, Sepehr Rezaee, Zeinab~Sadat Taghavi, Arad Maleki, Kian Shamsaie, Mahdi Hajialilue, et~al.,
\newblock ``Scanning trojaned models using out-of-distribution samples,''
\newblock in {\em The Thirty-eighth Annual Conference on Neural Information Processing Systems}.

\bibitem{moakhar2023seeking}
Arshia~Soltani Moakhar, Mohammad Azizmalayeri, Hossein Mirzaei, Mohammad~Taghi Manzuri, and Mohammad~Hossein Rohban,
\newblock ``Seeking next layer neurons' attention for error-backpropagation-like training in a multi-agent network framework,''
\newblock {\em arXiv preprint arXiv:2310.09952}, 2023.

\bibitem{taghavi2023change}
Zeinab~Sadat Taghavi, Ali Satvaty, and Hossein Sameti,
\newblock ``A change of heart: Improving speech emotion recognition through speech-to-text modality conversion,''
\newblock {\em arXiv preprint arXiv:2307.11584}, 2023.

\bibitem{rahimi-etal-2024-hallusafe}
Zahra Rahimi, Hamidreza Amirzadeh, Alireza Sohrabi, Zeinab Taghavi, and Hossein Sameti,
\newblock ``{H}allu{S}afe at {S}em{E}val-2024 task 6: An {NLI}-based approach to make {LLM}s safer by better detecting hallucinations and overgeneration mistakes,''
\newblock in {\em Proceedings of the 18th International Workshop on Semantic Evaluation (SemEval-2024)}, Atul~Kr. Ojha, A.~Seza Do{\u{g}}ru{\"o}z, Harish Tayyar~Madabushi, Giovanni Da~San~Martino, Sara Rosenthal, and Aiala Ros{\'a}, Eds., Mexico City, Mexico, June 2024, pp. 139--147, Association for Computational Linguistics.

\bibitem{taghavi2023imaginations}
Zeinab~Sadat Taghavi, Soroush Gooran, Seyed~Arshan Dalili, Hamidreza Amirzadeh, Mohammad~Jalal Nematbakhsh, and Hossein Sameti,
\newblock ``Imaginations of wall-e: Reconstructing experiences with an imagination-inspired module for advanced ai systems,''
\newblock {\em arXiv preprint arXiv:2308.10354}, 2023.

\bibitem{taghavi-etal-2023-ebhaam}
Zeinab Taghavi, Parsa~Haghighi Naeini, Mohammad~Ali Sadraei~Javaheri, Soroush Gooran, Ehsaneddin Asgari, Hamid~Reza Rabiee, and Hossein Sameti,
\newblock ``Ebhaam at {S}em{E}val-2023 task 1: A {CLIP}-based approach for comparing cross-modality and unimodality in visual word sense disambiguation,''
\newblock in {\em Proceedings of the 17th International Workshop on Semantic Evaluation (SemEval-2023)}, Atul~Kr. Ojha, A.~Seza Do{\u{g}}ru{\"o}z, Giovanni Da~San~Martino, Harish Tayyar~Madabushi, Ritesh Kumar, and Elisa Sartori, Eds., Toronto, Canada, July 2023, pp. 1960--1964, Association for Computational Linguistics.

\bibitem{taghavi2024backdooring}
ZeinabSadat Taghavi and Hossein Mirzaei,
\newblock ``Backdooring outlier detection methods: A novel attack approach,''
\newblock {\em arXiv preprint arXiv:2412.05010}, 2024.

\bibitem{rahimi-etal-2024-nimz}
Zahra Rahimi, Mohammad~Moein Shirzady, Zeinab Taghavi, and Hossein Sameti,
\newblock ``{NIMZ} at {S}em{E}val-2024 task 9: Evaluating methods in solving brainteasers defying commonsense,''
\newblock in {\em Proceedings of the 18th International Workshop on Semantic Evaluation (SemEval-2024)}, Atul~Kr. Ojha, A.~Seza Do{\u{g}}ru{\"o}z, Harish Tayyar~Madabushi, Giovanni Da~San~Martino, Sara Rosenthal, and Aiala Ros{\'a}, Eds., Mexico City, Mexico, June 2024, pp. 148--154, Association for Computational Linguistics.

\bibitem{ebrahimi2024sharif}
Seyedeh~Fatemeh Ebrahimi, Karim~Akhavan Azari, Amirmasoud Iravani, Arian Qazvini, Pouya Sadeghi, Zeinab~Sadat Taghavi, and Hossein Sameti,
\newblock ``Sharif-mgtd at semeval-2024 task 8: A transformer-based approach to detect machine generated text,''
\newblock {\em arXiv preprint arXiv:2407.11774}, 2024.

\bibitem{ebrahimi2024sharifa}
Seyedeh~Fatemeh Ebrahimi, Karim~Akhavan Azari, Amirmasoud Iravani, Hadi Alizadeh, Zeinab~Sadat Taghavi, and Hossein Sameti,
\newblock ``Sharif-str at semeval-2024 task 1: Transformer as a regression model for fine-grained scoring of textual semantic relations,''
\newblock {\em arXiv preprint arXiv:2407.12426}, 2024.

\bibitem{madry2017towards}
Aleksander Madry, Aleksandar Makelov, Ludwig Schmidt, Dimitris Tsipras, and Adrian Vladu,
\newblock ``Towards deep learning models resistant to adversarial attacks,''
\newblock {\em arXiv preprint arXiv:1706.06083}, 2017.

\bibitem{DBLP:journals/corr/abs-2007-08489}
Hadi Salman, Andrew Ilyas, Logan Engstrom, Ashish Kapoor, and Aleksander Madry,
\newblock ``Do adversarially robust imagenet models transfer better?,''
\newblock {\em CoRR}, vol. abs/2007.08489, 2020.

\bibitem{croce2020reliable}
Francesco Croce and Matthias Hein,
\newblock ``Reliable evaluation of adversarial robustness with an ensemble of diverse parameter-free attacks,''
\newblock in {\em International conference on machine learning}. PMLR, 2020, pp. 2206--2216.

\bibitem{felipe-campos-kitamura_2018}
Felipe~Campos Kitamura,
\newblock ``Head ct - hemorrhage,'' 2018.

\bibitem{brainmri}
Sartaj Bhuvaji, Ankita Kadam, Prajakta Bhumkar, Sameer Dedge, and Swati Kanchan,
\newblock ``Brain tumor classification (mri),'' 2020.

\bibitem{cohen2020covid}
Joseph~Paul Cohen, Paul Morrison, and Lan Dao,
\newblock ``Covid-19 image data collection,''
\newblock {\em arXiv}, 2020.

\bibitem{msoud-nickparvar_2021}
Msoud Nickparvar,
\newblock ``Brain tumor mri dataset,'' 2021.

\end{thebibliography}

\end{document}